%% file: main.tex
\documentclass[letterpaper, 10 pt, conference]{ieeeconf}  % Comment this line out if you need a4paper

\IEEEoverridecommandlockouts                              % This command is only needed if 
\usepackage{cite}
\usepackage{amsmath,amssymb,amsfonts}
\usepackage{algorithmic}
\usepackage{graphicx}
\usepackage{textcomp}
\usepackage{xcolor}
\usepackage{multirow}

\input{preambel/additional_packages.tex}

\title{\LARGE \bf
Bayesian Optimization-based Tire Parameter and Uncertainty Estimation for Real-World Data
}

\author{Sven Goblirsch$^{1}$, Benedikt Ruhland$^{1}$, Johannes Betz$^{2}$ and Markus Lienkamp$^{1}$ % <-this % stops a space
\thanks{*This work is a result of the joint research project ATLAS-L4. The project is supported by the German Federal Ministry for Economic Affairs and Climate Action (BMWK), based on a decision of the German Bundestag. The author is solely responsible for the content of this publication.} % <-this % stops a space
\thanks{$^{1}$S. Goblirsch, B. Ruhland and M. Lienkamp are with the Department of Mobility Systems Engineering, TUM School of Engineering and Design, Technical University of Munich, 85748 Garching, Germany. } % Munich Institute of Robotics and Machine Intelligence (MIRMI).}
\thanks{$^{2}$J. Betz is with the Professorship of Autonomous Vehicle Systems, TUM School of Engineering and Design, Technical University of Munich, 85748 Garching, Germany; Munich Institute of Robotics and Machine Intelligence (MIRMI). \newline{}
Corresponding author: \href{mailto:sven.goblirsch@tum.de}{sven.goblirsch@tum.de}}
}

\begin{document}

% Switch to single column mode
\onecolumn

% Manually insert the copyright notice outside of the two-column layout
\begin{center}
    \textcopyright \ 2025 IEEE. Personal use of this material is permitted. Permission from IEEE must be obtained for all other uses, including reprinting/republishing this material for advertising or promotional purposes, collecting new collected works for resale or redistribution to servers or lists, or reuse of any copyrighted component of this work in other works.
\end{center}

% Switch back to two-column mode
\twocolumn

\maketitle
\thispagestyle{empty}
\pagestyle{empty}

%%%%%%%%%%%%%%%%%%%%%%%%%%%%%%%%%%%%%%%%%%%%%%%%%%%%%%%%%%%%%%%%%%%%%%%%%%%%%%%%
\input{sections/00_abstract.tex}
\input{sections/01_introduction.tex}

\input{sections/02_relatedwork.tex}
\input{sections/03_methodology.tex}
\input{sections/04_results.tex}
\input{sections/05_conclusion.tex}

\section*{ACKNOWLEDGMENT}

As the first author, S. Goblirsch initiated and designed the paper's structure. He implemented the force calculation and parts of the Bayesian optimization and conducted the studies shown in this paper.
B. Ruhland implemented the sensitivity analysis and a first version of the Bayesian optimization.
J. Betz and M. Lienkamp made an essential contribution to the concept of the research project.
Both revised the paper critically for important intellectual content.
M. Lienkamp gives final approval for the version to be published and agrees to all aspects of the work.
As a guarantor, he accepts responsibility for the overall integrity of the paper.

\bibliographystyle{IEEEtran}
\bibliography{IEEEabrv, bibliography/literature.bib}

\end{document}

%% file: preambel/additional_packages.tex
\usepackage{tikz}
\usepackage{pgfplots}
\usepackage{float}

\pgfplotsset{compat=1.18}
\usepgfplotslibrary{statistics}

\usepackage[per-mode=fraction, detect-weight=true]{siunitx}
\usepackage[hidelinks]{hyperref}
\usepackage{caption}
\usepackage{subcaption}
\usepackage{multirow, graphicx}

\definecolor{TUMBlue}{RGB}{0,101,189}%      Pantone 300
\definecolor{TUMWhite}{RGB}{255,255,255}%
\definecolor{TUMBlack}{RGB}{0,0,0}%
\definecolor{TUMBlue1}{RGB}{0,51,89}%       Pantone 540
\definecolor{TUMBlue2}{RGB}{0,82,147}%      Pantone 301
\definecolor{TUMGray1}{RGB}{51,51,51}%
\definecolor{TUMGray2}{RGB}{127,127,127}%
\definecolor{TUMGray3}{RGB}{204,204,204}%
\definecolor{TUMBlue3}{RGB}{100,160,200}%   Pantone 542
\definecolor{TUMBlue4}{RGB}{152,198,234}%   Pantone 283
\definecolor{TUMIvory}{RGB}{218,215,203}%
\definecolor{TUMOrange}{RGB}{227,114,34}%
\definecolor{TUMGreen}{RGB}{162,173,0}%

%% file: sections/00_abstract.tex
\begin{abstract}

This work presents a methodology to estimate tire parameters and their uncertainty using a Bayesian optimization approach.
The literature mainly considers the estimation of tire parameters but lacks an evaluation of the parameter identification quality and the required slip ratios for an adequate model fit.
Therefore, we examine the use of Stochastical Variational Inference as a methodology to estimate both - the parameters and their uncertainties.
We evaluate the method compared to a state-of-the-art Nelder-Mead algorithm for theoretical and real-world application.
The theoretical study considers parameter fitting at different slip ratios to evaluate the required excitation for an adequate fitting of each parameter.
The results are compared to a sensitivity analysis for a Pacejka Magic Formula tire model.
We show the application of the algorithm on real-world data acquired during the Abu Dhabi Autonomous Racing League and highlight the uncertainties in identifying the curvature and shape parameters due to insufficient excitation.
The gathered insights can help assess the acquired data's limitations and instead utilize standardized parameters until higher slip ratios are captured.
We show that our proposed method can be used to assess the mean values and the uncertainties of tire model parameters in real-world conditions and derive actions for the tire modeling based on our simulative study.

\end{abstract}

%% file: sections/01_introduction.tex
\section{Introduction}
\label{sec:Introduction}

Accurate system identification is crucial for adequate performance in vehicle modeling and simulation. Utilizing a sophisticated vehicle model can, for instance, improve the side-slip angle estimation quality of online state estimators~\cite{chindamo_vehicle_2018, goblirsch_2024} and the control performance of trajectory tracking controllers in autonomous vehicles~\cite{raji_2023}. However, parameter mismatch can falsify the results of the incorporated models and yield insufficient performance.

While the vehicle dimensions and mass can easily be measured, identifying the tire parameters remains challenging due to the high nonlinearities and various environmental influences on the tire behavior~\cite{farroni_trip-id_2018}.
Generally, the approach used for tire model identification can be split into two groups: test bench usage and real-world testing. The first requires expensive testing routines for each tire and often does not accurately picture the tire-road surface interaction as either different shapes or materials are used. Furthermore, the operating condition of the tire can deviate from that of the actual vehicle~\cite{farroni_tricktireroad_2016, farroni_trip-id_2018}. The latter requires an additional tire force measurement, either done by measurement rims, trailers, or tire force estimators~\cite{acosta_2017}. This kind of data generally incorporates more noise on parameters and sensors, and usually, not the entire tire curve is captured due to limited excitation. To ensure adequate fitting, specific vehicle maneuvers must be included in the driving data~\cite{farroni_trip-id_2018, farroni_tricktireroad_2016}. However, performing stationary circular driving scenarios or other maneuvers requires large areas and dedicated test sessions, which are often inaccessible~\cite{dikici_2025}.

\begin{figure}[!t]
    \centering
    \vspace*{0.1cm}
    \includegraphics[width=\columnwidth]{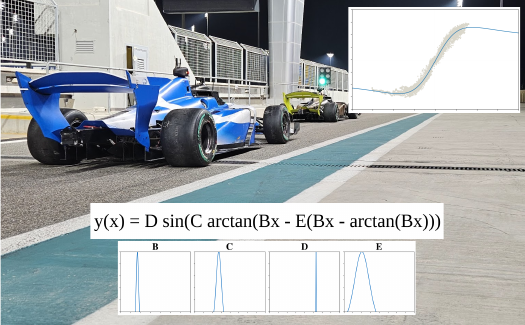}
    \caption{Resulting tire curve and parameter uncertainties for the front axle of a Dallara EAV24 race car.}
    \label{fig: sensor-setup}
\end{figure}

We propose a Bayesian-optimization-based parameter identification process to enable tire model parametrization for single-track models based on real-world data and assess the remaining uncertainties. The introduced method provides a better understanding of the tire model parameters and the required slip ratios for adequate system identification. Our main contributions are as follows:

\begin{itemize}
    \item We utilize Stochastical Variational Inference (SVI) for tire parameter and uncertainty estimation and evaluate the impact of excitation on the parameter accuracy. We evaluate the results using a simulative excitation study and real-world data recorded with an autonomous racecar.
    \item We perform a global sensitivity analysis of the Pacejka Magic Formula tire model and evaluate the influence of each parameter at different excitation levels. The results are compared to the SVI uncertainties.
    \item We provide an open-source parametrization tool, including data preprocessing and axle force calculation for single-track models. The tool features two optimization algorithms: Nelder-Mead and SVI for the Magic Formula Simple tire model. The code is available at \url{https://github.com/TUMFTM/Tire_Parameter_and_Uncertainty_Estimation}.
\end{itemize}

%% file: sections/02_relatedwork.tex
\section{Related Work}
\label{sec:RelatedWork}

Even though the Pacejka Magic Formula \cite{pacejka_tire_2006} was introduced decades ago, it is still one of the most common tire models. Therefore, many different approaches have been developed to identify the model parameters. Alagappan et al.~\cite{alagappan_comparison_2015} provide a comprehensive overview of different algorithms to optimize the Magic Formula parameters. One of the findings was that different combinations of the parameters could yield almost identical results due to the cross-correlations of the parameters. Rao et al.~\cite{rao_2006} perform a study to identify the relationship between the Magic Formula coefficients. They use a finite element model to model the physical behavior of the tire. The study underlines the impact of each parameter on the model performance. For instance, the shape factor is often set to a defined value for longitudinal and lateral models~\cite{bakker_1987}. However, an additional fitting highly improves the model's performance.  Furthermore, the impact of different environmental conditions is analyzed, showing that solely scaling the tire curve yields insufficient results~\cite{rao_2006}.

Kiebre et al.~\cite{kiebre_sensitivity_2011} perform a sensitivity analysis of the Magic Formula and the Fiala tire model and analyze the impact on the output force of the tire model. They show that the tire force is mainly influenced by the relative tire slip, the $D$ parameter, and the cornering stiffness. They do not distinguish between different excitation levels.
Albinson et al.~\cite{albinsson_2017} design a tire force excitation method to actively excite the tire for a friction estimator. Thereby, they perform a local sensitivity analysis for each parameter. Their analysis showed that the curvature parameter does not affect the linear tire region. Further, a minimum excitation is required to estimate the peak parameter $D$. Analog to \cite{rao_2006}, they achieve better results when estimating the shape parameter simultaneously.

Several authors mention the cross-correlation of the parameters~\cite{albinsson_2017, rao_2006, alagappan_comparison_2015, pacejka_tire_2006}. Therefore, different combinations can yield similar curves, making parameter identification more challenging. Still, utilizing fixed parameters can decrease the model performance, as shown in \cite{rao_2006, albinsson_2017}. Further, in the case of real driving data, different authors mention the importance of explicit driving maneuvers, and thus tire excitation, to facilitate sufficient model identification \cite{rao_2006, farroni_tricktireroad_2016, farroni_trip-id_2018, oosten_1992}. However, no analysis has been performed on how much excitation is required to adequately fit all Magic Formula tire model parameters.

New approaches for tire model fitting often utilize neural networks and show improvements for the considered area of tire excitation~\cite{dikici_2025, chrosniak_2024, olazagoitia_2020}. Dikici et al.~\cite{dikici_2025} highlight the importance of such methods to eliminate the use of specific maneuvers. While those methods yield good results for the acquired test data, different results might be attained when exceeding the excitations recorded in the test data. To avoid nonphysical parametrization, the network output is often bound to specific values~\cite{chrosniak_2024, olazagoitia_2020}. However, no excitation studies or parameter quality assessments have been performed. Such an investigation is, for instance, performed by Voser et al.~\cite{voser_2010} highlighting the importance of a peak in the curve to accurately estimate the peak parameter. However, such investigations have to be manually performed for each parameter.

Bayesian optimization offers a method to directly estimate the model's parameters and their uncertainties. An et al.~\cite{an_2024} use Bayesian optimization to estimate the damping and stiffness coefficients as well as mass, inertia, and wheelbase of a vertical vehicle model. Unjhawala et al.~\cite{unjhawala_2023} use a Sequential Monte Carlo method to estimate roll stiffness, damping coefficient, rolling resistance, and tire stiffness of a lateral vehicle model. However, no assessment of the tire nonlinearity is performed.
Wang et al.~\cite{wang_2023} utilize Bayesian optimization to optimize the model parameters of a single-track model. However, the tire parameter identification yielded worse results than the full model identification. They provide no insights into the utilized tire excitation or the required driving data. Boyali et al.~\cite{boyali_2021} identify the cornering stiffness and the center of gravity position with Bayesian optimization and analyze the uncertainty for front and rear cornering stiffness. An excitation study and identification of the full tire model were not performed. Also, the study was conducted only on simulation data.
Lionti et al.~\cite{lionti_2024} perform a parameter estimation based on the Approximate Bayes Computation Sequential Monte Carlo method to estimate the tire parameters and their uncertainties of a lateral vehicle dynamics model. However, they do not analyze the uncertainties compared to an excitation study and only show the approach based on simulation data.

%% file: sections/03_methodology.tex
\section{Methodology}
\label{sec:Methodology}

\subsection{Magic Formula Tire Model}

We conduct our studies based on the Magic Formula Simple model~\cite{pacejka_tire_2006}. The model describes the longitudinal and lateral tire force response to certain excitations in longitudinal (slip ratio in percent) and lateral (slip angle in rad) directions. Thereby, an empirical equation consisting of 4 parameters is used. The $B$ parameter describes the tire stiffness, $C$ is the shape of the curve, $D$ is the peak, and $E$ is used to model the curvature. Furthermore, $S_h$ and $S_v$ can be used to shift the curve in the vertical and horizontal directions to account, for instance, for ply steer and conicity~\cite{pacejka_tire_2006, bakker_1987}. The resulting tire force $Y$ based on the excitation $X$ is calculated as follows.

\begin{equation}
    y(x) = D\sin\left(C\arctan\left(Bx-E\left(Bx-\arctan(Bx)\right)\right)\right)
\end{equation}
\begin{equation}
    x = X + S_h
\end{equation}
\begin{equation}
    Y(X) = y(x) + S_v
\end{equation}

The developers of this formula recommend using a $C$ parameter of 1.65 for longitudinal and 1.3 for lateral tire models. Furthermore, the $D$ parameter should range between 0.7 and 1.3 for dry road conditions. For racing tires, this value can go up to 2~\cite{bakker_1987, pacejka_tire_2006}.

\subsection{Bayesian Optimization Parameter Fitting}

Fitting those parameters can be done with various algorithms, as shown in~\cite{alagappan_comparison_2015}. However, none of these presented algorithms outputs a parameter uncertainty.

We, therefore, utilize SVI to assess both mean value and uncertainty of each parameter. 
The principle of SVI is combining variational inference with stochastic optimization~\cite{hoffman_2012}.
Compared to the Markov-Chain-Monte-Carlo methods, this method requires less tuning whilst being able to handle larger datasets. 

A joint probability distribution $p(z, x)$ with $x$ observations and $z$ latent variables is used. $p(z)$ resembles the prior and $p(z|x)$ the likelihood~\cite{hoffman_2012, blei_2017}. For our studies, we use the baseline implementation of~\cite{bingham_2019}.

\begin{equation}
    p_\theta(z, x) = \frac{p(x|z)}{p(z)}
\end{equation}

We define the latent variables $z$ as a multivariate normal distribution with the tire model parameters $B$, $C$, $D$, and $E$ as the mean vector and their cross-correlations as the covariance matrix. 
The observation $x$ is the calculated tire force based on the Magic Formula with the current model parameters in $z$. To account for noise in the sensor data, we model the observations as a normal distribution with an additional parameter $\sigma$ for the variance. 
These observations are compared to the actual tire forces to optimize the likelihood function and get better estimates of $z$. We use the evidence lower bound (ELBO) function as the objective function to compute the divergence between prior $p(z)$ and posterior $q(z)$.

\begin{equation}
    ELBO(q) = \mathbb{E} \left[ \log p(z, x) \right] - \mathbb{E} \left[ \log q(z) \right]
\end{equation}

To ensure physical behavior of the derived model, the tire parameters are bound to the following values.

\begin{table}[ht]
    \caption{Bounds of the Magic Formula Parameters in the SVI algorithm.}
    \begin{center}
    \bgroup
    \def\arraystretch{1.2}
    \begin{tabular}{|c|c|c|c|}
        \hline
         Parameter & Minimum & Maximum \\ \hline
         B & 5.0 & 40.0 \\
         C & 1.0 & 3.0  \\
         D & 0.1 & 2.0  \\
         E & -1.0 & 1.0  \\ \hline
    \end{tabular}
    \egroup
    \label{tab:params_opt}
\end{center}
\end{table}

\subsection{Sensitivity Study}

The SVI algorithm outputs a mean vector and a covariance for the Magic Formula parameters. These can be used to analyze the model quality based on the input data and, thus, the contained excitation level. To evaluate the results of this method, we use a sensitivity study to assess the impact of each parameter on the output curve. 

One common approach to evaluate the impact of different parameters on the model output is calculating the Sobol Index~\cite{sobol_2001}. The total variance of a function $y=f(x)$ with $x$ being a set of model parameters and y being the scalar model output is calculated by using variance decomposition. Thereby, $V_i$ resembles the contribution of the parameter $i$ and $V_{ij}$ the contribution of the interaction between two parameters. $V_{1,...,p}$ accounts for higher influence due to interaction between multiple parameters.

\begin{equation}
    V(y) = \sum_{i=1}^p V_i + \sum_{i=1}^{p-1} \sum_{j=i+1}^p V_{ij}+...+V_{1,...,p}
\end{equation}

The Total Sobol Index $S_{T_i}$ then measures the total effect of each parameter and its interactions on the output of $y$. For our analysis, we only consider the first and second-order influences and, therefore, neglect $V_{1,...,p}$.

\begin{equation}
    S_{T_i} = \frac{V_i + \sum_{j \neq i} V_{ij}}{V (Y)}
\end{equation}

\subsection{Tire Parameter Fitting}

To evaluate the SVI approach on real-world data and, thus, incorporate parameter scatter and sensor noise, we calculate the axle forces similar to~\cite{farroni_tricktireroad_2016}. The longitudinal force on the center of gravity $F_{x, cog}$ is calculated based on the mass of the vehicle $m$, the longitudinal acceleration $a_x$, the resulting drag force $F_{drag}$ and the rolling resistance of the vehicle $F_{roll}$. This force is split to front $F_{x, f}$ and rear axle $F_{x,r}$ considering the vertical axle loads $F_{z, f}$ and $F_{z, r}$ for deceleration maneuvers. Since a rear-wheel-drive drivetrain is considered, the entire longitudinal force is obtained by the rear axle for acceleration scenarios. The vertical load calculation considers longitudinal load transfer due to acceleration as well as aerodynamic lift forces on the front and rear axle.

The axle lateral forces $F_{y, f}$ and $F_{y, r}$ are calculated based on the lateral acceleration $a_y$, the distance of the center of gravity to the rear axle $l_r$, the wheelbase $l$, the yaw moment of inertia $I_{zz}$ and the derivate of the yaw rate $r$. Further, we extend the approach by adding the torque of a limited-slip differential $T_{LSD}$ calculated based on the preload, coast, and drive settings of the differential~\cite{gadola_2018}. The drive torque is derived from the longitudinal force of the rear axle and clipped to the engine brake torque for deceleration maneuvers.

Calculating the axle forces in real-world scenarios enables us to examine the excitations reached and underlines the importance of parameter uncertainty for real-world scenarios. 
\begin{equation}
    F_{x, cog} = m a_x + F_{drag} + F_{roll}
\end{equation}
\begin{equation}
    F_{x, f} =
    \begin{cases} 
        0 & \text{if } F_{x, cog} > 0 \\
        F_{x, cog} \frac{F_{z, f}}{F_{z, f} + F_{z, r}} & \text{else}
    \end{cases}
\end{equation}
\begin{equation}
    F_{x,r} = F_{x, cog} - F_{x,f}
\end{equation}
\begin{equation}
    F_{y,f} = \frac{l_r m a_y - I_{zz} \left( \frac{dr}{dt} \right) + T_{LSD}}{l}
\end{equation}
\begin{equation}
    F_{y, r} = m a_y - F_{y, f}
\end{equation}

The lateral forces of the front axle are further transformed by the steering angle $\delta$ into the tire coordinate frame~\cite{bechtloff_schatzung_2018}.

\begin{equation}
    F_{y, f, tf} = \frac{1}{\cos \delta} (F_{y, f} - F_{x, f} \sin \delta)
\end{equation}

The axle states $\lambda_i$ (longitudinal slip) and $\alpha_i$ (slip angle) are calculated as follows, with $v_{x, i}$ and $v_{y, i}$ being the velocities at the front and rear axle, transformed by the distance to the center of gravity and the steering angle. $\omega_{left, i}$ and $\omega_{right, i}$ are the measured rotational wheel speeds.

\begin{equation}
    \lambda_i = \frac{\frac{\omega_{left, i}+\omega_{right, i}}{2} r_{dyn, i} - v_{x, i}}{v_{x, i}} 
\end{equation}
\begin{equation}
    \alpha_i = - \arctan \left( \frac{v_{y, i}}{v_{x,i }} \right)
\end{equation}

We extend the slip calculation by the dynamic tire radius $r_{dyn, i}$ similar to~\cite{schabauer_2020}. $r_{0, i}$ is the unloaded tire radius, $r_{s, i}$ the static tire radius, and $r_{i}$ is the unloaded radius of the non-rotating tire. $c_{i}$ is the global tire stiffness and $d_{r, i}$ the tire speed expansion factor.

\begin{equation}
    r_{dyn, i} = r_{0, i} \cdot \frac{2}{3} + r_{s, i} \cdot \frac{1}{3}
\end{equation}
\begin{equation}
    r_{0, i} = r_{i} + d_{r, i} \left( \frac{\omega_{left, i} + \omega_{right, i}}{2} \right) ^2
\end{equation}
\begin{equation}
    r_{s, i} = r_{i} - \frac{F_z, i}{2c_{i}}
\end{equation}

To account for sensor misalignments, we perform an offset calculation as presented in~\cite{farroni_tricktireroad_2016}. We further prefilter the sensor signals for excessive noise and potential outliers. The published module contains three filter options: Moving Average, Savitzky-Golay, and Gaussian filter. For our studies, we solely utilize the Savitky-Golay filter. Furthermore, gear shifts are filtered as the spiking moment in the drivetrain causes a twist in the axle, resulting in spikes in the wheel speed encoder signal.

To account for different ply steer or conicity of left and right tire, asymmetrical setup, or other impacts yielding a shift in the axle forces, we calculate $S_v$ and $S_h$ beforehand. These parameters solely require data in the linear tire area as they resemble the shift of the origin.

To ensure adequate data coverage and prevent overfitting of the linear region, we prefilter the data using a nearest-neighbor approach. This way we ensure an even data distribution over the collected tire excitations.

\subsection{Experiments}

We first perform an analysis based on simulation data to evaluate the impact of different excitation levels on the tire parameter identification procedure. This enables full coverage of all slip states with the same data density and noise level. We create a Magic Formula curve with the parameters in Tab.~\ref{tab:params_matmodel}. We further apply noise to the slip data and the final force output to provide a more realistic setting. Both optimizers use the parameter bounds shown in Tab.~\ref{tab:params_opt}. To assess the impact of different excitation levels, we create noisy data with a maximum slip value corresponding to the excitation level. Note that we use the same quantity of data points for each excitation level to neglect the impact of data density. We interpret the excitation as a longitudinal slip ratio to provide a better understanding compared to the maximum level. A slip ratio of 1.0 indicates full sliding of the tire. Afterward, we optimize the tire parameters using the previously described SVI algorithm for each excitation level and compare the results with a state-of-the-art Nelder-Mead algorithm.

\begin{table}[H]
    \caption{Parameters of the mathematical model used for the simulative analysis.}
    \begin{center}
    \bgroup
    \def\arraystretch{1.2}
    \begin{tabular}{|c|c|c|c|}
        \hline
         B & C & D & E \\ \hline
         15.0 & 2.0  & 1.5  & 0.8 \\ \hline
    \end{tabular}
    \egroup
    \label{tab:params_matmodel}
\end{center}
\end{table}

To evaluate the results, we perform a sensitivity analysis. We vary the parameters by 10 percent and calculate the total Sobol index. We focus on this region to analyze each parameter's impact on this respective curve. Considering different initial parameters yields different results. Therefore, no global excitation requirements can be derived. This finding aligns with the implications on parameter cross-correlations mentioned by~\cite{albinsson_2017, rao_2006, alagappan_comparison_2015, pacejka_tire_2006}. Nevertheless, this sensitivity analysis provides insights into the variation of the curve at different slip ratios. It can be used to interpret the results of the optimized parameters and the calculated uncertainty by the SVI approach.

Even though the mathematical model can be used to evaluate the general applicability of the SVI algorithm and assess the excitation requirements in an optimal setting, real-world data features additional challenges. These contain non-normal-distributed signal noise, scattered parameters, and deviations from the mathematical model. We use real-world data recorded during the Abu Dhabi Autonomous Racing League to evaluate the introduced SVI-based tire parameter optimization. The vehicle side slip angle was measured using a Kistler SF-Motion Correvit sensor at 500~Hz. The accelerations and yaw rate were measured with a Vectornav VN310 at 800~Hz. All other signals were recorded at a frequency of 100~Hz. The data is first filtered with the settings shown in Tab.~\ref{tab:real_filter} and down-sampled to 100~Hz to account for the different sensor frequencies and varying noise levels. Further, offsets of the sensors are compensated as previously described.

\begin{table}[!h]
    \caption{Settings used for preprocessing the raw sensor data.}
    \begin{center}
    \bgroup
    \def\arraystretch{1.2}
    \begin{tabular}{|c|c|c|c|}
        \hline
         Sensor & Freq. & Filter & Settings \\ \hline
         Correvit & 500~Hz & Savitzky-Golay  & Window: 200, Order: 3 \\ \hline
         IMU & 800~Hz & Savitzky-Golay  & Window: 500, Order: 5 \\ \hline
         CAN & 100~Hz & Savitzky-Golay & Window: 30, Order: 3 \\ \hline
    \end{tabular}
    \egroup
    \label{tab:real_filter}
\end{center}
\end{table}

%% file: sections/04_results.tex
\section{Results}
\label{sec:Results}

\subsection{Excitation Study}

\begin{figure}[!h]
    \centering
    \vspace*{0.1cm}
    \includegraphics[width=\columnwidth]{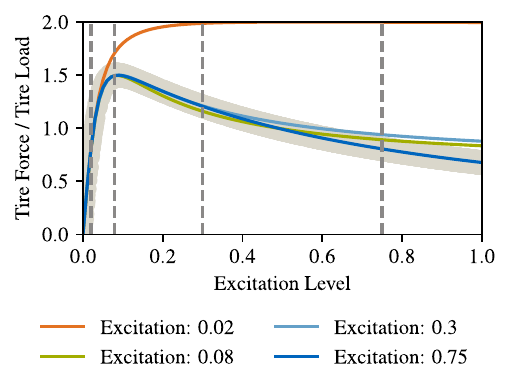}
    \caption{Tire model fit for different excitation levels.}
    \label{fig:matmodel_tirecurve}
\end{figure}

First, the proposed estimation of parameters and uncertainties is evaluated in a simulative setting. Fig.~\ref{fig:matmodel_tirecurve} shows the resulting curve with additional noise. Four curves are displayed for models fitted at different excitation levels. The first maximum slip ratio was defined at 2\%, located in the curve's linear area. The resulting tire curve highly overshoots the peak. The excitation level at 8\% covers the peak of the curve. However, none of the sliding region is covered. Consequently, the curve fits well until this slip ratio but deviates at pure sliding. Only the curve with 75\% excitation covers the entire force curve of the tire.

These results match the findings of \cite{acosta_2017}, showing that enough excitation has to be covered to estimate the maximum of the tire curve. Note that we evaluate a full factorial parametrization of the Magic Formula. Approaches exist to estimate this value solely from low slip conditions. However, their performance is still questioned in the literature~\cite{acosta_2017}. Further, high excitation is needed to model the pure sliding area accurately.  

\begin{figure}
    \centering
    \vspace*{0.1cm}
    \includegraphics[width=\columnwidth]{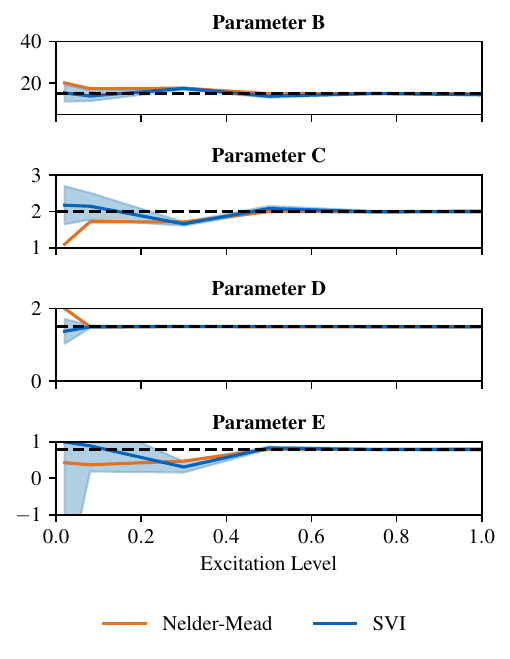}
    \caption{Parameter fitting and uncertainty at different excitation levels. The black dotted line visualizes the reference parameters. The light blue area visualizes the doubled standard deviation.}
    \label{fig:matmodel_paramhist}
\end{figure}

Fig.~\ref{fig:matmodel_paramhist} shows the history of the estimation of the Magic Formula parameters for different excitation levels. Both algorithms - Nelder-Mead and SVI - show high deviations in the linear region. Once the peak of the curve is covered, both algorithms accurately estimate the peak parameter $D$. Until then, the SVI approach outputs a higher standard deviation, implying a less accurate fit. The other parameters cannot be accurately estimated until 50\% of the maximum is reached. The $E$ parameter especially shows a high estimated standard deviation.

\begin{figure}
    \centering
    \vspace*{0.1cm}
    \includegraphics[width=\columnwidth]{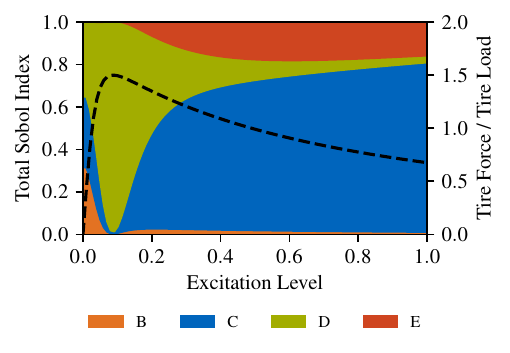}
    \caption{Sobol Index over slip ratio. The black dotted line shows the underlying tire force curve.}
    \label{fig:sobol-matmodel}
\end{figure}

We perform a sensitivity analysis to evaluate the impact of each parameter at different slip ratios. As shown in Fig.~\ref{fig:sobol-matmodel}, the initial stiffness is equally affected by $B$, $C$, and $D$. This matches the literature as the tire stiffness can also be expressed as the product $BCD$. Once the peak of the curve is reached, the $D$ parameter resembles the main impact factor. The parameter $E$ barely affects the initial shape of the curve but is responsible for the sliding region. This matches the SVI results with a higher uncertainty of the $E$ parameter until enough data is covered.

\subsection{Real-World Data}

\begin{figure}
    \centering
    \vspace*{0.1cm}
    \includegraphics[width=\columnwidth]{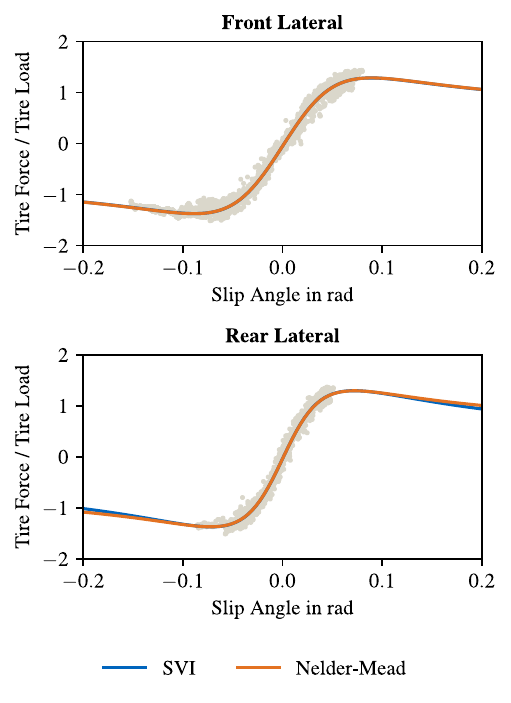}
    \caption{Lateral tire curves of the Dallara EAV24 vehicle on the Yas Marina Circuit.}
    \label{fig: real-force}
\end{figure}

Fig.~\ref{fig: real-force} shows the final tire curves acquired in three runs of TUM Autonomous Motorsport~\cite{betz2023tum} during the Abu Dhabi Autonomous Racing League. The plot shows a fit for the introduced SVI and the benchmark Nelder-Mead algorithm. As can be seen, both can accurately fit the tire curve. However, the SVI algorithm slightly deviates from the Nelder-Mead result at the rear axle for higher slip angles.

The front axle experienced higher slip angles due to the car's understeering behavior. Even though relatively high slip angles are reached on the front axle, the rear does not exceed the maximum tire limit. Consequently, fitting the $C$ parameter of the tire model should be more accurate on the front axle. Further, no pure sliding occurred on both axles. Thus, a fitting of the E parameter cannot be performed for both axles based on our previous findings.

\begin{figure}
    \centering
    \vspace*{0.1cm}
    \includegraphics[width=\columnwidth]{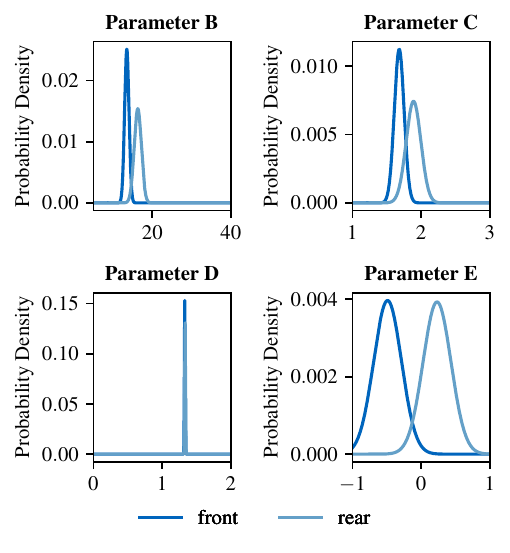}
    \caption{Parameter uncertainties of the Magic Formula tire model for the shown real-world data.}
    \label{fig: real-bell}
\end{figure}

The probability densities of the SVI fitting are depicted in Fig.~\ref{fig: real-bell}. For the $E$ parameter, both front and rear axles show a high spread in the distribution. The $D$ parameter shows a good fit in both cases, which inlines with the tire curves, where a peak can be seen. The $B$ and $C$ parameters show a better fit for the front axle, where higher slip values occurred.

As a consequence of these results, a fixed value of $E$ should be considered - either according to the literature or based on test bench data. Once more data in higher slip regimes have been collected, refitting the entire tire curve would be recommended. Further, the SVI method provides insights into the slip areas where the model can be confidently applied. Based on the fitting results, an application in pure sliding areas would not be recommended.

%% file: sections/05_conclusion.tex
\section{Conclusion}
\label{sec:conclusion}

In this paper, we suggested a Bayesian-optimization-based optimization scheme to quantify not only the parameters of a Magic Formula tire model but also their uncertainties. We demonstrated the uncertainty estimation in a theoretical study. The results were compared with a sensitivity analysis highlighting the impact of each parameter at different excitations. We further demonstrated the application to real-world data. The paper provides new insights into the excitation requirements for adequate model parametrization. These can be used to evaluate model limitations for certain slip ratios. To ensure physical behavior of the model, standardized parameters can be employed until sufficient data is acquired. The method can be extended to more sophisticated tire models in the future. Furthermore, the current study is limited to one vehicle and one set of tires. A survey for different road conditions or vehicle types remains future work.